\documentclass[letterpaper, 10 pt, conference]{ieeeconf}  

\IEEEoverridecommandlockouts                              

\usepackage{amsmath}
\usepackage{amssymb}
\usepackage{mathtools}

\usepackage[utf8]{inputenc} 
\usepackage[T1]{fontenc}    
\usepackage{hyperref}       
\usepackage{url}            
\usepackage{booktabs}       
\usepackage{amsfonts}       
\usepackage{nicefrac}       
\usepackage{microtype}      
\usepackage{xcolor}         

\usepackage{graphicx}
\usepackage{epsfig}
\usepackage{algorithmic}
\usepackage{multirow,bigdelim}
\usepackage{subfigure}
\usepackage{soul}
\usepackage[framemethod=tikz]{mdframed}
\usepackage{todonotes}
\usepackage{balance}

\newtheorem{prom}{Problem}

\newtheorem{asm}{Assumption}
\newtheorem{thm}{Theorem}

\newtheorem{lemma}{Lemma}
\newtheorem{remark}{Remark}

\title{\LARGE \bf
Physical Deep Reinforcement Learning Towards Safety Guarantee
}

\author{Hongpeng Cao$^{1}$, Yanbing Mao$^{2}$, Lui Sha$^{3}$, Marco Caccamo$^{1,4}$ 
\thanks{$^{1}$ Hongpeng Cao is with School of engineering and design, Technical University of Munich, Munich, 85748, Germany
        {\tt\small cao.hongpeng@tum.de}}%
\thanks{$^{2}$ Yanbing Mao is with Engineering Technology Division, Wayne State University, Detroit, MI 48201, USA
        {\tt\small hm9062@wayne.edu}}%
\thanks{$^{3}$ Lui Sha is with Department of Computer Science, University of Illinois at Urbana-Champaign, Urbana, IL 61801, USA {\tt\small lrs@illinois.edu}}%
\thanks{$^{1, 4}$ Marco Caccamo is with School of Engineering and Design, Technical University of Munich (TUM), Munich, Germany and Munich Institute of Robotics and Machine Intelligence, Technical University of Munich (TUM), Munich, Germany. {\tt\small mcaccamo@tum.de}
}
}

\begin{document}

\maketitle
\thispagestyle{empty}
\pagestyle{empty}

\begin{abstract}
Deep reinforcement learning (DRL) has achieved tremendous success in many complex decision-making tasks of autonomous systems with high-dimensional state and/or action spaces. However, the safety and stability still remain major concerns that hinder the applications of DRL to safety-critical autonomous systems. To address the concerns, we proposed the Phy-DRL: a physical deep reinforcement learning framework. The Phy-DRL is novel in two architectural designs: i) Lyapunov-like reward, and ii) residual control (i.e., integration of physics-model-based control and data-driven control). The concurrent physical reward and residual control empower the Phy-DRL the (mathematically) provable safety and stability guarantees. Through experiments on the inverted pendulum, we show that the Phy-DRL features guaranteed safety and stability and enhanced robustness, while offering remarkably accelerated training and enlarged reward.  
\end{abstract}

\section{Introduction}
Over the past decades, reinforcement learning (RL) has demonstrated breakthroughs for sequential decision making in broad areas,  ranging from autonomous driving \cite{kendall2019learning} and finance \cite{abe2010optimizing} to chemical processes \cite{savage2021model} and games \cite{silver2018general}. These success of RL were inherently limited to fairly low-dimensional problems, i.e., the previous RL frameworks lacked scalability. To remove the limitation, deep reinforcement learning (DRL) arises, which relies on deep neural networks for the powerful function approximation and representation learning of action value function, action policy, environment states, to name a few \cite{mnih2015human, silver2016mastering}. DRL has achieved tremendous success in many complex decision making tasks with high-dimensional state and action spaces, such as vision-based control of robots \cite{levine2016end}. DRL thus holds a promise for revolutionizing the artificial intelligence (AI) towards a higher-level understanding of the visual world, with tangible industrial and economic impact. But the recent frequent incidents of AI-assisted autonomous systems overshadow the revolutionizing potential of DRL as well, especially for the safety-critical systems. For instance, according to statistics released by the US National Highway Traffic Safety Administration (NHTSA) in the year 2022,  automakers reported nearly 400 crashes linked to self-driving, driver-assist technologies in 11 months \cite{NHTSA}. Unfortunately, NHTSA recently added 11 new deaths to a growing list of fatalities tied to the use of (semi)-automated driving system \cite{NHTSA11}. Hence, the particularly safety enhanced DRL is even more vital today, which aligns well with the market’s need for reliable deep learning technologies and motivates the research of safety and stability of DRL-enabled autonomous system, including the training and inference \cite{wachi2020safe,brunke2022safe,bucsoniu2018reinforcement}.

Generally, a stable system shall have a property that, if the system starts from a safe region, it will eventually converge to the goal state, known as \textit{asymptotically stable} \cite{drazin1992nonlinear}.
The safety guarantee is thus desirable to prompt reliable DRL. To do so, the control Lyapunov function (CLF) is to proposed to encode such property into the reward function, such that the learning agent is regulated to learn to stabilize the system. For example, Chang and Gao in \cite{chang2021stabilizing} proposed to learn a Lyapunov function from sampled data and use it as an additional critic network to regulate the control policy optimization toward the decrease of the Lyapunov critic values. The challenge moving forward is how to design DRL to exhibit (mathematically) provable stability guarantee.  Importantly, the seminal study in \cite{westenbroek2022lyapunov} discovered that if the reward of DRL is CLF-like, the stability of DRL-enabled autonomous systems is mathematically guaranteed.

Alternatively, CLF can be used to constrain the exploration state into the safety set, such that all actions will lead the system to decent on defined CLF~\cite{perkins2002lyapunov}, i.e., towards being stable.~\cite{berkenkamp2017safe} proposes to use Lyapunov stability theory to define a safety set and get the agent only explore and learn policy in the safety set. It also shows that the safety set can be expanded by using the Gaussian process to learn the dynamics. ~\cite{perkins2002lyapunov} used Lyapunov to construct basic-level control laws that can enjoy safety and performance guarantees. A reinforcement learning agent is introduced to switch the low-level control laws to finish the task optimally. Similarly, ~\cite{qin2022quantifying, wachi2020safe, fisac2018general, cheng2019end} propose to use prior knowledge of the model to constrain the exploration state with desired safety specifications, and the DRL is only allowed to explore in the constrained space to ensure safety. Typically, the safe region is derived by analyzing the stability of a linearized model and is often conservative, limiting the performance of the learned policy. Therefore, those approaches need to construct a more accurate dynamic model from the interaction data to expand the safe region~\cite{berkenkamp2017safe, cheng2019end}. In addition, the safety architectures~\cite{sha2001using,alshiekh2018safe,hewing2020learning} proposed in the control community can also be employed to ensure system-level safety for DRL-enabled systems for both training and inference. However, these architectures normally do not make assumptions about the internals of the learning agent, and thus can not encourage stability or safety during learning. Moreover, the safety envelope developed in those architectures might be limited by under-modeling errors presented in linearized models.

Although the success, the challenges moving forward are 
\begin{itemize}
    \item What is formal guidance of constructing control Lyapunov function for reward of DRL? 
    \item How to design DRL to have provable concurrent safety and stability guarantees? 
\end{itemize}

To address the challegnes, we proposed the physical deep reinforcement learning framework (Phy-DRL). The novelty of Phy-DRL is twofold, which can be summarized as follows. 
\begin{itemize}
    \item \textit{Physics-Model-Regulated Reward}, which provides guidance of constructing safety- and stability-aware reward.  
    \item \textit{Residual Control}, an integration of physics-model-based control and data-driven control, which in conjunction with regulated reward empower the Phy-DRL the (mathematically) provable safety and stability guarantees. 
\end{itemize}

This paper is organized as follows. In Section II, we present

\section{Preliminaries}
For convenience, Table \ref{notation}  summarizes the notations used throughout the paper. 
\begin{table}[ht] {
\centering
\caption{Table of Notation}
\begin{tabular}{|l|}
\hline
$\mathbb{R}^{n}$:~the set of $\emph{n}$-dimensional real vectors  \\ \hline $\mathbb{N}$:~the set of natural numbers           \\ \hline
$[\mathbf{x}]_{i}$:~the $i$-th entry of vector $\mathbf{x}$       \\ \hline
$[\mathbf{W}]_{i,:}$:~ the $i$-th row of matrix $\mathbf{W}$ \\ \hline $[\mathbf{W}]_{i,j}$:~ the element at row $i$ and column $j$ of matrix $\mathbf{W}$\\ \hline
$\mathbf{P} \succ 0$:~ the matrix $\mathbf{P}$ is positive definite \\ \hline $\top$:~the matrix or vector transposition  \\ \hline
$\mathbf{I}_{n}$:~  the the $n \times n$-dimensional identity matrix  \\ \hline $\mathbf{1}_{n}$:~ $n$-dimensional vector of all ones \\ \hline
\end{tabular}\label{notation}}
\end{table}

\subsection{Real Plant}
Without loss of generality, the real system is described by 
\begin{align}
\mathbf{s}(k+1) = \mathbf{A}\mathbf{s}(k) + \mathbf{B}\mathbf{a}(k) + \mathbf{f}(\mathbf{s}(k), \mathbf{a}(k)), ~~~~k \in \mathbb{N}  \label{realsys}
\end{align}
where $\mathbf{s}(k) \in \mathbb{R}^{n}$ is the real-time system state, $f(\mathbf{s}(k), \mathbf{a}(k)) \in \mathbb{R}^{n}$ is the unknown model mismatch, $\mathbf{a}(k) \in \mathbb{R}^{m}$ is the control command. 

The considered safety problems stem from practical regulations or constraints on system states, which motives the following safety set.
\begin{align}
\textbf{Safe Set:}  ~&{\mathbb{X}} \triangleq \left\{ {\left. {\mathbf{s} \in {\mathbb{R}^n}} \right|\underline{\mathbf{v}} \le {\mathbf{D}} \cdot \mathbf{s} - \mathbf{v} \le \overline{\mathbf{v}} }\right\}, \label{aset2}
\end{align}
where $\mathbf{D}$, $\mathbf{v}$, $\overline{\mathbf{v}}$ and $\underline{\mathbf{v}}$ are given in advance. 

\begin{remark}[Safety Problem Example]
The condition in \eqref{aset2} can cover a significant number of safety problems that are due to operation regulations and/or safety constraints. Taking the autonomous vehicles driving in school zone in Winter as one example \cite{mao2023sL1}, according to traffic regulation, the vehicle speed shall be around 15 mph, while to prevent slipping and sliding for safe driving in icy roads, the vehicle slip shall not be larger  than 4 mph. Given the information of regulation and safety constraint, we can let 
\begin{align}
\mathbf{s} = \!\left[\!\! \begin{array}{l}
v\\
w
\end{array} \!\!\right]\!, \mathbf{D} = \!\left[\!\! {\begin{array}{*{20}{c}}
1\!&\!0\\
1\!&\!{ - r}
\end{array}} \!\!\right],  \mathbf{v} = \!\left[\!\! \begin{array}{l}
15\\
0
\end{array} \!\!\right]\!, \overline{\mathbf{v}} = \!\left[\!\! \begin{array}{l}
2\\
4
\end{array} \!\!\right]\!, \underline{\mathbf{v}} = \!\left[\!\! \begin{array}{l}
-2\\
-4
\end{array} \!\!\right]\!, \nonumber
\end{align}
such that condition in \eqref{aset2} can be equivalently transformed to 
\begin{align}
&-2 \le v - 15 \le 2, \label{con1}\\
&-4 \le v - r \cdot w \le 4, \label{con2}
\end{align}
where $v$, $w$ and $r$ denote vehicle's longitudinal velocity, angular velocity and wheel radius, respectively. The inequality \eqref{con1} means the maximum allowable difference with traffic regulated velocity (i.e., 15 mph) is 2 mph. While the inequality \eqref{con2} means the vehicle slip (defined as $v - r \cdot w$) is constrained to be not larger than 4 mph.   
\end{remark}

\begin{figure*}[t]
	\centering
	\subfigure{\includegraphics[width=0.8\linewidth]{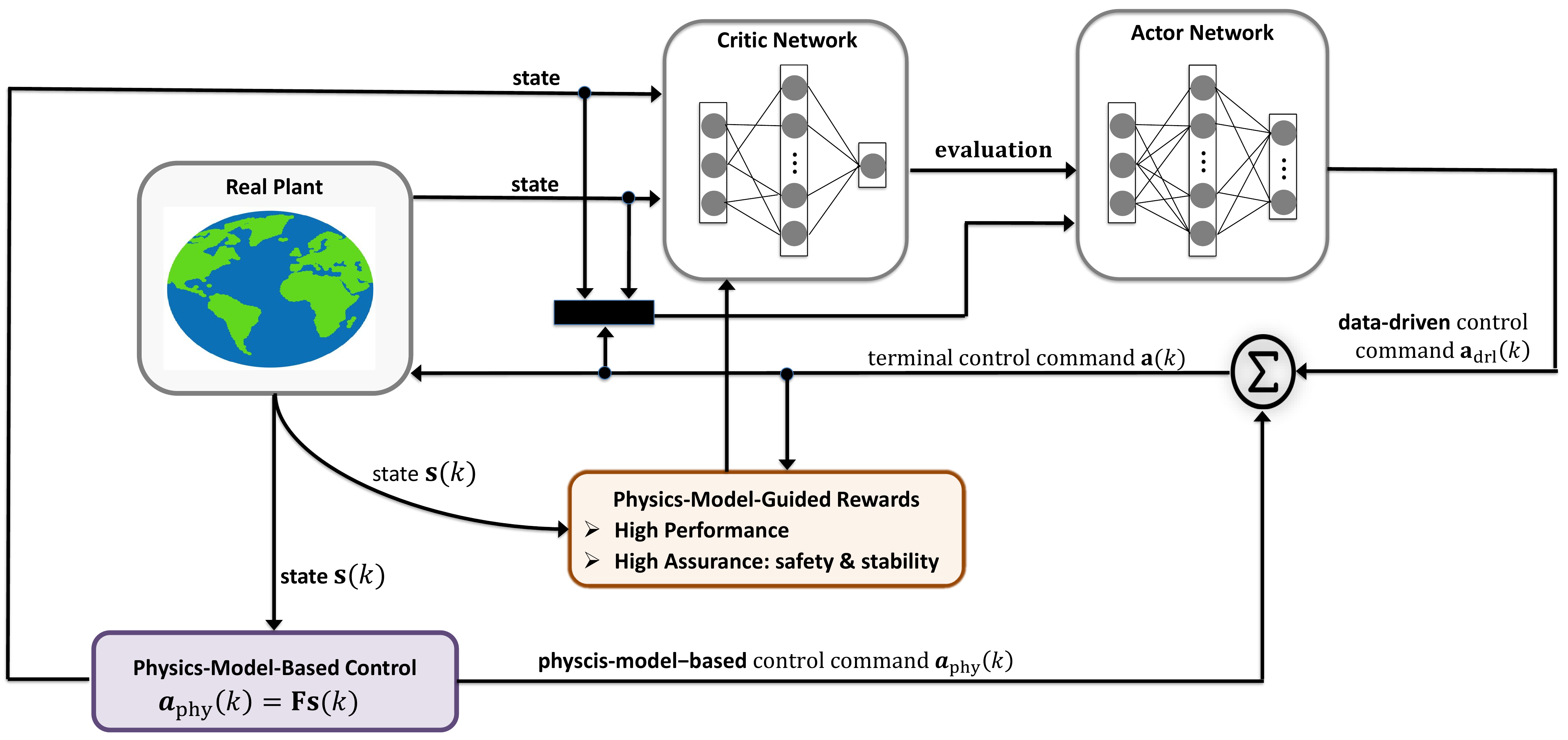}\hspace{-0.35cm}}
	\caption{The plot shows the diagram of the proposed Phy-DRL framework It consists of a real plant, a physics-model-based controller, a DRL algorithm of \textit{actor-critic} architecture, and a physics-model-guided reward module.}
        \label{prdrl}
\end{figure*}

\subsection{Deep Reinforcement Learning}
In this paper, we propose the deep reinforcement learning for generating the high-performance control command  $\mathbf{a}(k)$ for real plant \eqref{realsys}. The developed DRL is based on the deep deterministic policy gradient \cite{lillicrap2015continuous}, which learn a deterministic control policy $\pi$ that maximizes the expected return from the initial state distribution: 
\begin{align}
\!\!\!{Q^\pi}( {\mathbf{s}(k), \mathbf{a}(k)})
& = {\mathbf{E}_{\mathbf{s}(k) \sim \mathbb{S}}}\left[ {\sum\limits_{t = k}^N {{\gamma^{t-k}}  \mathcal{R}\left( {\mathbf{s}(t),\mathbf{a}(t)} \right)}} \right] \nonumber \\
& = {\mathbf{E}_{\mathbf{s}(k) \sim \mathbb{S}}}\left[ {\sum\limits_{t = k}^N {{\gamma ^{t-k}}  \mathcal{R}\left( {\mathbf{s}(t),\pi \left( {\mathbf{s}(t)} \right)} \right)} } \right]\!\!,  \label{bellman}
\end{align}
where $\mathbb{S}$ represents the state space, $\mathcal{R}(\cdot)$ maps a state-action-next-state triple to a real-valued reward, $\gamma \in [0,1]$ is the discount factor, controlling the relative importance of immediate and future rewards. 

\subsection{Problem Formulation}
The matrices $\mathbf{A}$ and $\mathbf{B}$ in the dynamics model \eqref{realsys} represent the available physical knowledge pertaining to the real plant. The investigated problem will be based on the available knowledge, which is formally stated below. 
\begin{prom}
How to leverage the available system matrix $\mathbf{A}$ and control structure matrix $\mathbf{B}$ pertaining to the real plant \eqref{realsys} to design DRL towards safety and stability guarantees? \label{prob}
\end{prom}

The proposed solution to the Problem \ref{prob} is the physical DRL, whose framework is shown in Figure \ref{prdrl}. The Phy-DRL has two  architectural innovations: i) physics-model-regulated reward, and ii) residual control, i.e., integration of physics-model-based control and data-driven control from DRL.

\section{Phy-DRL: Physics-Model-Regulated Reward}
The investigation of physics-model-regulated reward aims at guidance of constructing safety- and stability-aware reward, which is indispensable for answering the Problem \ref{prob}. 

\subsection{Safety Envelope}
The current safety set formula \eqref{aset2} is not ready for developing the safety- and stability-aware reward $\mathcal{R}(\cdot)$ in \eqref{bellman}. To move forward, we introduce an equivalent variant of safety set \eqref{aset2}: 
\begin{align}
\widehat{\mathbb{X}} \triangleq \left\{ {\left. {\mathbf{s} \in {\mathbb{R}^n}} \right| -\mathbf{1}_{h} \le  {\mathbf{d}} \le \underline{\mathbf{D}} \cdot \mathbf{s}, ~~
\overline{\mathbf{D}} \cdot \mathbf{s} \le \mathbf{1}_{h}}\right\}, \label{set2} 
\end{align}
where for $i \in \{1,2,\ldots, h\}$, 
\begin{align}
 [{\mathbf{d}}]_{i} &\triangleq \begin{cases}
		1, & \text{if}~\left[\underline{\mathbf{v}} +  \mathbf{v}\right]_{i} > 0 \\
        1, &\text{if}~\left[\overline{\mathbf{v}} +  \mathbf{v}\right]_{i} < 0 \\
		-1, & \text{if}~\left[\underline{\mathbf{v}} +  \mathbf{v}\right]_{i} > 0,~\left[\underline{\mathbf{v}} +  \mathbf{v}\right]_{i} < 0
	\end{cases} \label{compbadd3} 
\end{align} 
with the $\overline{\mathbf{v}}$, $\underline{\mathbf{v}}$ and ${\mathbf{v}}$ given in \eqref{aset2}, and the subscript $h \in \mathbb{N}$ indicating the number of constraint or regulation conditions.  

The two sets $\widehat{\mathbb{X}}$ and ${\mathbb{X}}$ can be equivalent, which is formally stated in the following lemma, whose proof appears in Appendix \ref{apreenvelope}. 
\begin{lemma}
Consider the set ${\mathbb{X}}$ defined in \eqref{aset2} and the set $\widehat{\mathbb{X}}$ defined in \eqref{set2}. The ${\mathbb{X}} = \widehat{{\mathbb{X}}}$ holds, if and only if 
$\overline{\mathbf{D}} = \frac{{\mathbf{D}}}{\overline{\Lambda}}$ and $\underline{\mathbf{D}} = \frac{{\mathbf{D}}}{\underline{\Lambda}}$, where for $i,j \in \{1,2,\ldots, h\}$, 
\begin{align}
[\overline{\Lambda}]_{i,j} &\triangleq \begin{cases}
		0, & \text{if}~i \ne j\\
		\left[\overline{\mathbf{v}} +  \mathbf{v}\right]_{i}, & \text{if}~\left[\underline{\mathbf{v}} +  \mathbf{v}\right]_{i} > 0 \\
        \left[\underline{\mathbf{v}} +  \mathbf{v}\right]_{i}, & \text{if}~\left[\overline{\mathbf{v}} +  \mathbf{v}\right]_{i} < 0 \\
		\left[\overline{\mathbf{v}} +  \mathbf{v}\right]_{i}, & \text{if}~\left[\overline{\mathbf{v}} +  \mathbf{v}\right]_{i} > 0,~\left[\underline{\mathbf{v}} +  \mathbf{v}\right]_{i} < 0
	\end{cases} \label{compbadd1} \\
 [\underline{\Lambda}]_{i,j} &\triangleq \begin{cases}
		0, \!\!& \text{if}~i \ne j\\
		\left[\underline{\mathbf{v}} +  \mathbf{v}\right]_{i}, \!\!& \text{if}~\left[\underline{\mathbf{v}} +  \mathbf{v}\right]_{i} > 0 \\
        \left[\overline{\mathbf{v}} +  \mathbf{v}\right]_{i}, \!\!& \text{if}~\left[\overline{\mathbf{v}} +  \mathbf{v}\right]_{i} < 0 \\
		\left[-\underline{\mathbf{v}} -  \mathbf{v}\right]_{i}, \!\!& \text{if}~\left[\overline{\mathbf{v}} +  \mathbf{v}\right]_{i} > 0,~\left[\underline{\mathbf{v}} +  \mathbf{v}\right]_{i} < 0.
	\end{cases}\label{compbadd2} 
\end{align} \label{preenvelope}
 \end{lemma}

We now introduce the safety envelope, which constitutes a building block of safety- and stability-aware reward. 
\begin{align}
{\Omega} &\buildrel \Delta \over = \left\{ {\left. {\mathbf{s} \in {\mathbb{R}^n}} \right|{\mathbf{s}^\top}{\mathbf{P}}\mathbf{s} \le 1,~{\mathbf{P} } \succ 0} \right\}. \label{set3}
\end{align}

The following lemma builds a connection between the safety envelope ${\Omega}$ and the safety set $\widehat{\mathbb{X}}$. Specifically, it provides a condition under which the safety envelope   ${\Omega}$ is a subset of safety set $\widehat{\mathbb{X}}$.
\begin{lemma} 
Consider the safety set $\widehat{\mathbb{X}}$ and the safety envelope $\Omega$ defined in \eqref{set2} and \eqref{set3}, respectively. The ${\Omega} \subseteq \widehat{\mathbb{X}}$ holds, if 
\begin{align}
&{\overline{\mathbf{D}}}{\mathbf{P}^{-1}}\overline{\mathbf{D}}^\top \le \mathbf{I}_{h} ~\text{and} \nonumber\\
& {{\left[ {{\underline{\mathbf{D}}}\mathbf{P}^{ - 1}\underline{\mathbf{D}}^\top} \right]}_{i,i}} \!=\! \begin{cases}
		\ge 1, \!\!\!\!& \text{if}~[{\mathbf{d}}]_i = 1\\
		\le 1, \!\!\!\!& \text{if}~[{\mathbf{d}}]_i = -1
	\end{cases}\!, i \in \{1, \ldots, h\}. \label{resc}
 \end{align} \label{envelope}
\end{lemma}
\begin{remark}[Usage of Lemma \ref{envelope}]
 The condition \eqref{resc} in  Lemma \ref{envelope} will be used to compute the model-based control commands (see LMIs \eqref{th4} and \eqref{th4a}). 
\end{remark}

\subsection{Safety- and Stability-Aware Reward}
In light of the condition of safety envelope \eqref{set3}, we are ready to propose safety- and stability-aware reward. For the sake of simplifying the remaining presentations, we define: 
\begin{align}
\overline{\mathbf{A}} \buildrel \Delta \over = \mathbf{A} + {\mathbf{B}} {\mathbf{F}}, \label{asysma}
\end{align}
where $\mathbf{F}$ is a design matrix, whose computation is presented in section IV. Hereto, the proposed reward is 
\begin{align}
\mathcal{R}( {\mathbf{s}(k),\mathbf{a}(k)}) &= \left[ {\mathbf{s}^\top(k){{\overline{\mathbf{A}}^\top}\mathbf{P}\overline{\mathbf{A}} }\mathbf{s}(k)}  - {\mathbf{s}^\top(k\!+\!1) }\mathbf{P}\mathbf{s}(k\!+\!1) \right]  \nonumber\\
&\hspace{3.0cm} + g( \mathbf{s}(k),\mathbf{a}(k)), \label{reward}
\end{align}
where $\mathbf{P}$ is given in \eqref{set3},  the term $g( \mathbf{s}(k),\mathbf{a}(k))$ is for high operation performance (such as avoiding jerk for comfortable driving), while remaining terms are motivated by the aim of safety and stability guarantees. 

\begin{remark}
The matrices $\mathbf{F}$ and $\mathbf{P}$ in the reward formula are computed based on the physics-model knowledge represented by $\mathbf{A}$ and $\mathbf{B}$. The computations of $\mathbf{F}$ and $\mathbf{P}$ will be presented in the next section, which are subject to LMIs \eqref{th1}--\eqref{th4a}. 
\end{remark}

\section{Phy-DRL: Residual Control}
As shown in Figure \ref{prdrl},  the terminal control command $\mathbf{a}(k)$ from Phy-DRL is given in the residual form: 
\begin{align}
\mathbf{a}(k) = \mathbf{a}_{\text{drl}}(k) + \mathbf{a}_{\text{phy}}(k),
\label{residual}
\end{align}
where $\mathbf{a}_{\text{drl}}(k)$ denotes the date-driven control command from DRL, while $\mathbf{a}_{\text{phy}}(k)$ denotes the physics-model-based control command computed according to 
\begin{align}
\mathbf{a}_{\text{phy}}(k) = \mathbf{F} \mathbf{s}(k),  ~\text{with}~ \mathbf{F} = \mathbf{R}\mathbf{Q}^{-1}, \mathbf{Q}^{-1} = \mathbf{P}. 
\label{modelbased}
\end{align}
The matrices $\mathbf{A}$, $\mathbf{B}$, $\mathbf{Q}^{-1} = \mathbf{P}$,  $\mathbf{F}$ and $\mathbf{R}$ in \eqref{reward} and \eqref{modelbased} are computed through solving
the following LMIs via LMI toolbox \cite{boyd1994linear}:   
\begin{align}
&\left[\!\! {\begin{array}{*{20}{c}}
\alpha \mathbf{Q} & \mathbf{Q}\mathbf{A}^\top + \mathbf{R}^\top\mathbf{B}^\top\\
\mathbf{A}\mathbf{Q} + \mathbf{B}\mathbf{R} & \mathbf{Q}
\end{array}} \!\!\right] \succ 0, \label{th1}\\
& \mathbf{I}_{h} - \overline{\mathbf{D}}{\mathbf{Q}}\overline{\mathbf{D}}^\top\succ 0,  \label{th4}\\
&{{\left[ {{\underline{\mathbf{D}}}\mathbf{Q}\underline{\mathbf{D}}^\top} \right]}_{i,i}} = \begin{cases}
		\ge 1, & [{\mathbf{d}}]_i = 1\\
		\le 1, & [{\mathbf{d}}]_i = -1
	\end{cases}, i \in \{1,\ldots, h\}\label{th4a}
\end{align}
where $\mathbf{d}$ is given in \eqref{compbadd3}, and $0 < \alpha < 1$ is a given scalar. 

We next present a property of real plant with the residual control, which will be used to prove the safety and stability guarantees of Phy-DRL in the next section.
\begin{lemma} 
For the systems \eqref{realsys} with residual control \eqref{residual}, define the function: 
\begin{align}
V(\mathbf{s}(k)) \buildrel \Delta \over = {\mathbf{s}^\top(k)} \cdot {{\mathbf{P}}} \cdot \mathbf{s}(k). \label{tradlya}
\end{align}
If the model-based control \eqref{modelbased} in the residual control \eqref{residual} satisfies the condition \eqref{th1}, the function $V(\mathbf{s}(k))$ along real plant satisfies 
\begin{align}
\!\!\!\!V( {{\mathbf{s}(k\!+\!1)}}) \!-\! V( {{\mathbf{s}(k)}}) \!=\! r(\mathbf{s}(k), \mathbf{a}(k)) \!+\! (\alpha \!-\! 1)  V( {{\mathbf{s}(k)}}) \label{formre}
\end{align}
where
\begin{align}
& r(\mathbf{s}(k), \mathbf{a}(k)) \nonumber\\
&\buildrel \Delta \over =   {{\left( \mathbf{f}(\mathbf{s}(k), \mathbf{a}(k))  \!+\!  \mathbf{B}\mathbf{a}_{\text{drl}}(k) \right)}^\top}\mathbf{P}(\mathbf{f}(\mathbf{s}(k), \mathbf{a}(k))  \!+\! \mathbf{B}\mathbf{a}_{\text{drl}}(k)) \nonumber\\
&~~~~~~~ + 2{{( {{\overline{\mathbf{A}}}{\mathbf{s}(k)}})^\top}}\mathbf{P} \left(\mathbf{f}(\mathbf{s}(k), \mathbf{a}(k)) \!+\!
  \mathbf{B}\mathbf{a}_{\text{drl}}(k)\right). \label{tradlyb}
\end{align} \label{fflema}
\end{lemma}

\section{Phy-DRL: Provable Safety and Stability Guarantees}
The conjunctive physics-model-regulated reward \eqref{reward} and 
residual control \eqref{residual} empowers the trained Phy-DRL the provable safety and stability guarantees. Before presenting the result, we introduce a practical assumption pertaining to the data-driven term \eqref{tradlyb}. 
\begin{asm} 
Along the real plant under the control of Phy-DRL, the function \eqref{tradlyb} satisfies 
\begin{align}
r(\mathbf{s}(k), \mathbf{a}(k)) < \beta(\mathbf{s}(k)). \label{asm1eq}
\end{align} \label{asm1}
\end{asm}

\begin{remark}
The upper bound $\beta(\mathbf{s}(k))$ in \eqref{asm1eq} is a function of system state $\mathbf{s}(k)$ only is motivated by the fact that both model-based control $\mathbf{a}_{\text{phy}}(k)$  and data-driven control $\mathbf{a}_{\text{drl}}(k)$ depend on system state only. 
\end{remark}

The safety and stability of Phy-DRL is formally presented in the following theorem. 
\begin{thm}
Consider the real plant \eqref{realsys} under control of Phy-DRL, whose reward is given in \eqref{reward} and control command is given in \eqref{residual} with \eqref{modelbased}, where the involved matrices $\mathbf{A}$, $\mathbf{B}$, $\mathbf{Q}^{-1} = \mathbf{P}$,  $\mathbf{F}$ and $\mathbf{R}$ satisfy the conditions \eqref{th1}--\eqref{th4a}. Under Assumption \ref{asm1},   
\begin{itemize}
    \item If $\frac{\beta(\mathbf{s}(k))}{1-\alpha} < 1$ holds for any $k \in \mathbb{N}$, the control policy of Phy-DRL renders the given safety envelope $\Omega$ \eqref{set3} invariant, i.e., if $\mathbf{s}(1) \in \Omega \subseteq \widehat{\mathbb{X}} = {\mathbb{X}}$, then $\mathbf{s}(k) \in \Omega \subseteq \widehat{\mathbb{X}} = {\mathbb{X}}$ for any $k \in \mathbb{N}$. 
    \item If $\beta(\mathbf{s}(k)) + (\alpha - 1) \cdot V\left( {{\mathbf{s}(k)}} \right) < 0$ holds for any $k \in \mathbb{N}$, the control policy of Phy-DRL asymptotically stabilizes the real system \eqref{realsys} and renders the given safety $\Omega$ \eqref{set3} invariant. 
\end{itemize}\label{thm2}
\end{thm}

For simplifying the explanation of reward \eqref{reward} and proof of Theorem \ref{thm2}, the real plant under the control of Phy-DRL is rewritten as 
\begin{align}
\!\!\!\!\mathbf{s}(k+1) = \overline{\mathbf{A}} \mathbf{s}(k) + {\mathbf{B}} \mathbf{a}_{\text{drl}}(k) + \mathbf{f}(\mathbf{s}(k),\mathbf{a}(k)), k \in \mathbb{N} \label{auxphydrl}  
\end{align}
where $\overline{\mathbf{A}}$ is defined in \eqref{asysma}. 

\begin{remark} [Reward Motivation and Explanation]
In light of \eqref{auxphydrl}, we obtain from \eqref{tradlyb} that 
\begin{align}
&r(\mathbf{s}(k), \mathbf{a}(k)) \nonumber\\
&= 2{{\left( {\overline{\mathbf{A}}{\mathbf{s}(k)}} \right)}^\top}\mathbf{P} \left(\mathbf{s}(k+1) - \overline{\mathbf{A}} \mathbf{s}(k)\right) \nonumber\\
&\hspace{1.8cm} + {{\left(\mathbf{s}(k+1) - \overline{\mathbf{A}} \mathbf{s}(k)\right)^\top)}}\mathbf{P}\left(\mathbf{s}(k+1) - \overline{\mathbf{A}} \mathbf{s}(k)\right) \nonumber\\
& = {{\left(\mathbf{s}(k+1) \right)}^\top}\mathbf{P}\left(\mathbf{s}(k+1)\right) - {\mathbf{s}^{\top}(k)\left( {{{\overline{\mathbf{A}}}^\top}\mathbf{P}\overline{\mathbf{A}}} \right){\mathbf{s}(k)}}, \nonumber
\end{align} 
which means the reward \eqref{reward} includes a sub-reward term $-r(\mathbf{s}_k, \mathbf{u}_k) = {\mathbf{s}_k^\top\left( {{{\overline{\mathbf{A}}}^\top}\mathbf{P}\overline{\mathbf{A}}} \right){\mathbf{s}_k}} - {{\left(\mathbf{s}_{k + 1} \right)}^\top}\mathbf{P}\left(\mathbf{s}_{k + 1}\right)$ that the data-driven control commands $\mathbf{a}_{\text{drl}}(k)$ from Phy-DRL try to maximize. We conclude that the reward \eqref{reward} has one objective of encouraging choices of control commands for decreasing $r(\mathbf{s}_k, \mathbf{u}_k)$ over time, such that  $r(\mathbf{s}_k, \mathbf{u}_k)$ can have a minimum upper-bound $\beta(\mathbf{s}(k))$ given in \eqref{asm1eq}. \label{ssremark}
\end{remark}

\begin{remark} [System Behavior and Phy-DRL Evaluation]
The results presented in Theorem \ref{thm2} can be used to evaluate the safety and stability of a trained Phy-DRL. Specifically, if  $\frac{\beta(\mathbf{s}(k))}{1-\alpha} < 1$, only system safety can be guaranteed, i.e., the stability cannot be guaranteed. One example of system behavior in this scenario is shown in phase plot in Figure \ref{ee} (a), where system states always stay inside the safety envelope (can be oscillating), but do not converge to the equilibrium. If $\beta(\mathbf{s}(k)) + (\alpha - 1) \cdot V\left( {{\mathbf{s}(k)}} \right) < 0$, both safety and stability can be guaranteed. The corresponding behavior in this scenario is depicted in Figure \ref{ee} (b). 
\begin{figure}[http]
    \centering
    \includegraphics[width=1\linewidth]{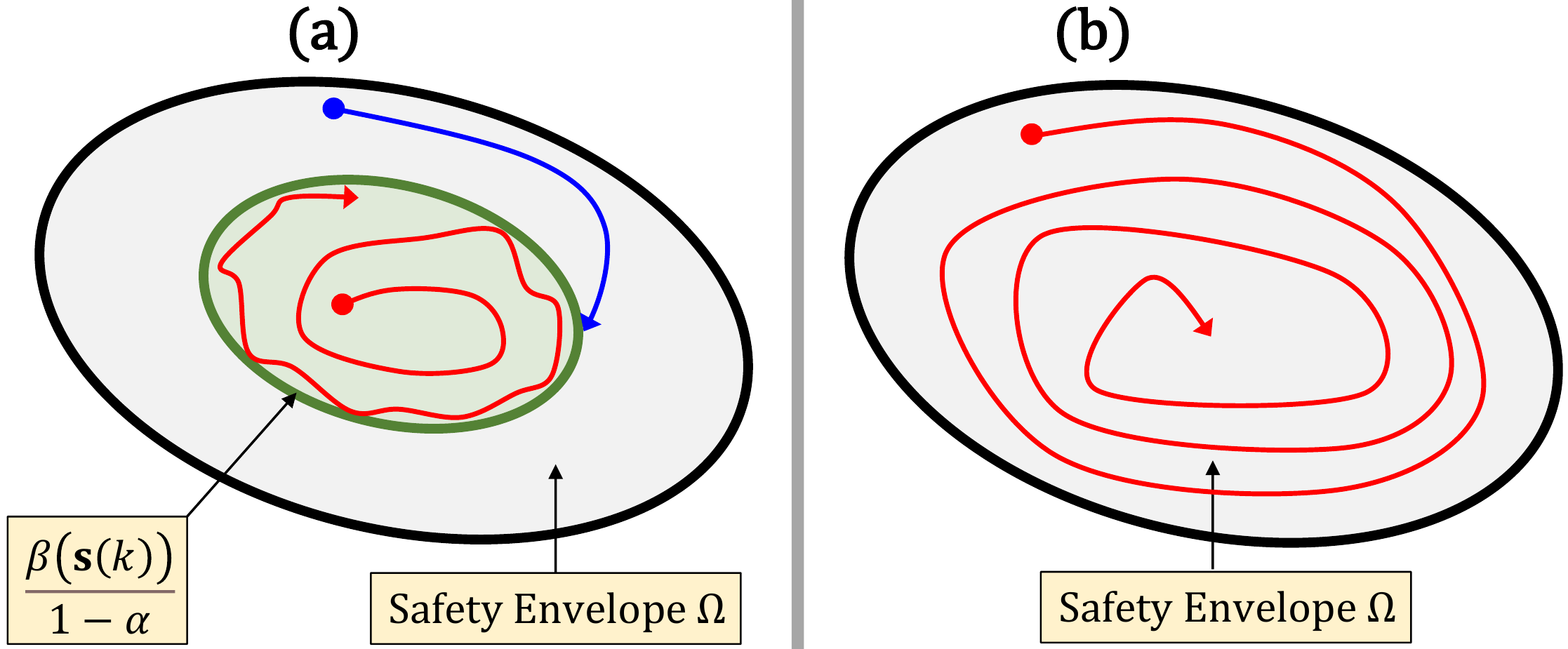}
    \caption{(a) Only safety is guaranteed, (b) both safety and stability are guaranteed. Dots denote the initial conditions (starting points).}
    \label{ee}
   \end{figure}
\label{evaluation}
\end{remark}

\begin{remark} [Residual Model Mismatch Learning]
The knowledge of $\beta(\mathbf{s}(k))$ is critical in safety and stability evaluation of Phy-DRL, discussed in Remark \ref{evaluation}. According to \eqref{tradlyb} and \eqref{asm1eq}, the $\beta(\mathbf{s}(k))$ can be obtained through learning the residual model mismatch $\mathbf{f}(\mathbf{s}(k),\mathbf{a}(k))$. Furthermore, according to \eqref{auxphydrl}, the mismatch $\mathbf{f}(\mathbf{s}(k),\mathbf{a}(k))$ can be learned from the samples $(\mathbf{s}(k+1), \mathbf{s}(k))$ (generated by the real plant under control of Phy-DRL), since the $\overline{\mathbf{A}}$, $\mathbf{B}$ and $\mathbf{a}_{\text{drl}}(k)$ in \eqref{auxphydrl} are known. 
\end{remark}

\section{Experiments}\label{sec:exp}
\label{expav}

\begin{figure}[htbp]
	\centering
	\subfigure{
	\includegraphics[width=0.75\columnwidth]{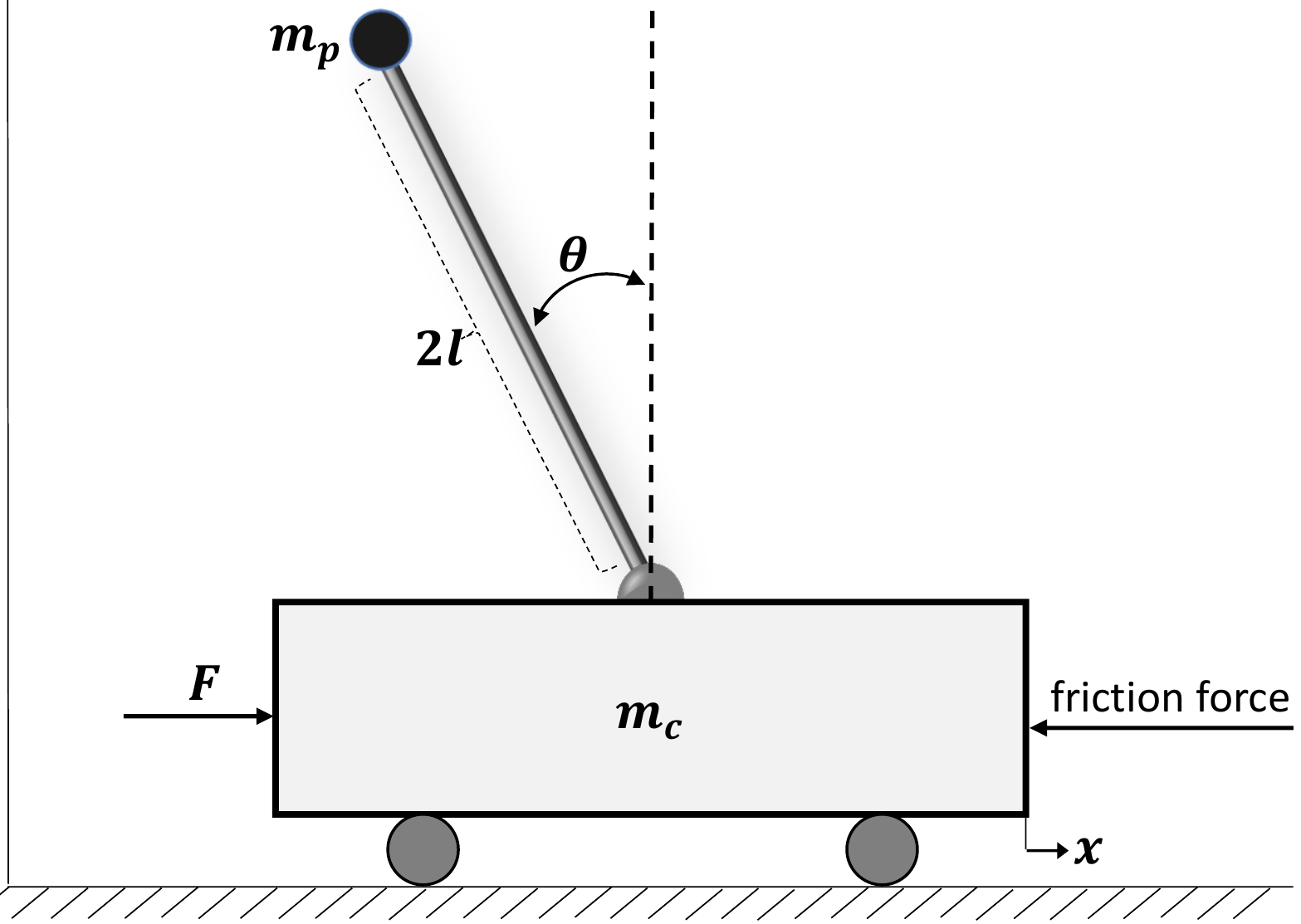}\hspace{-0.35cm}
	}
	\caption{Mechanical analog of inverted pendulums.}
        \label{df}
\end{figure}

We demonstrate the proposed Phy-DRL in an inverted pendulum case study, whose mechanical analog is shown in Figure~\ref{df}. The inverted pendulum system is characterized by the angle of the pendulum from vertical $\theta$, angular velocity of $\omega  \buildrel \Delta \over = \dot \theta$, the position of the cart $x$ and cart velocity $v \buildrel \Delta \over = \dot x$. The control goal is to stabilize the pendulum at the equilibrium $\mathbf{s}^* = [x^{*},v^{*},\theta^{*},\omega^{*}]^\top = [0,0,0,0]^\top$.  

To demonstrate the robustness of Phy-DRL, the following system matrix $\mathbf{A}$ and control structure matrix $\mathbf{B}$ are obtained without considering friction force, while the real plant is subject to friction force.  Specifically, we first take the dynamic model of the inverted pendulum described in~\cite{florian2007correct} and linearize it around the equilibrium $[x^{*},v^{*},\theta^{*},\omega^{*},] = [0,0,0,0]$, for which we assume that the obtained subsystem stays within a small neighborhood of this equilibrium, and we use the approximations: $\cos \theta  \approx 1$, $\sin \theta  \approx \theta$ and ${\omega ^2}\sin \theta  \approx 0$. 
\begin{align} 
\mathbf{A} &= \left[{\begin{array}{*{20}{c}}
1&{0.0333}&0&0\\
0&1&{ - 0.0565}&0\\
0&0&1&{0.0333}\\
0&0&{0.8980}&1
\end{array}} \right], \label{PM1}\\
\mathbf{B} &= \left[
0~~0.0334~~0~~- 0.0783
\right]^\top.  \label{PM2} 
\end{align}
The considered safety conditions are 
\begin{align} 
-0.6 \le x \le 0.6, ~~~-0.4 \le \theta < 0.4. \label{safetycond}
\end{align}
We let $\alpha = 0.8$. The matrices $\mathbf{P}$ and $\mathbf{F}$ are solved from LMIs \eqref{th1}--\eqref{th4a} via Matlab LMI toolbox:
\begin{align} 
&\mathbf{P} = \left[ {\begin{array}{*{20}{c}}
{{\rm{2}}{\rm{.0120}}}&{{\rm{0}}{\rm{.2701}}}&{{\rm{1}}{\rm{.4192}}}&{{\rm{0}}{\rm{.2765}}}\\
{{\rm{0}}{\rm{.2701}}}&{{\rm{2}}{\rm{.2738}}}&{{\rm{5}}{\rm{.1795}}}&{{\rm{1}}{\rm{.0674}}}\\
{{\rm{1}}{\rm{.4192}}}&{{\rm{5}}{\rm{.1795}}}&{{\rm{31}}{\rm{.9812}}}&{{\rm{4}}{\rm{.9798}}}\\
{{\rm{0}}{\rm{.2765}}}&{{\rm{1}}{\rm{.0674}}}&{{\rm{4}}{\rm{.9798}}}&{{\rm{1}}{\rm{.0298}}}
\end{array}} \right], \label{PM3}\\
&\mathbf{F} = \left[ {\begin{array}{*{20}{c}}
{0.7400}&{3.6033}&{35.3534}&{6.9982}
\end{array}} \right]. \label{PM4}
\end{align}

We let the high-performance reward $g( \mathbf{s}(k),\mathbf{a}(k)) = -a^2(k)$. Given the sub-reward and the knowledge \eqref{PM1}--\eqref{PM2}, the residual control \eqref{residual} and reward \eqref{reward} of Phy-DRL can be formed. 

\begin{figure}[ht]
\centering
\subfigure{\includegraphics[width=0.99\columnwidth]{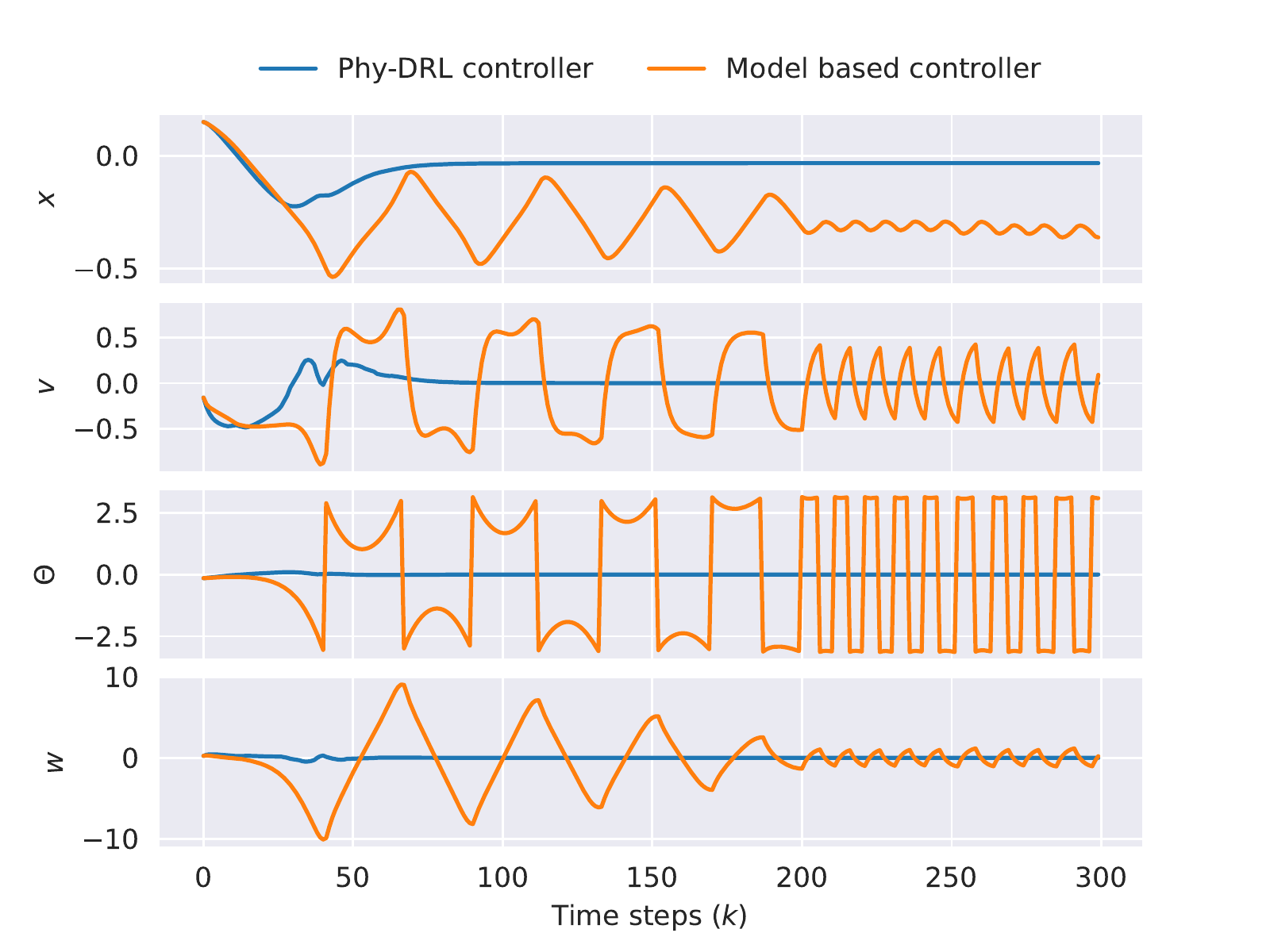}}
\caption{The plot shows an example of state trajectories of the system controlled by the proposed Phy-DRL controller and model-based controller.}
\label{trajectory}
\end{figure}

\begin{figure}[ht]
\centering   
\subfigure[Cost during training]{\label{fig:a}\includegraphics[width=78mm]{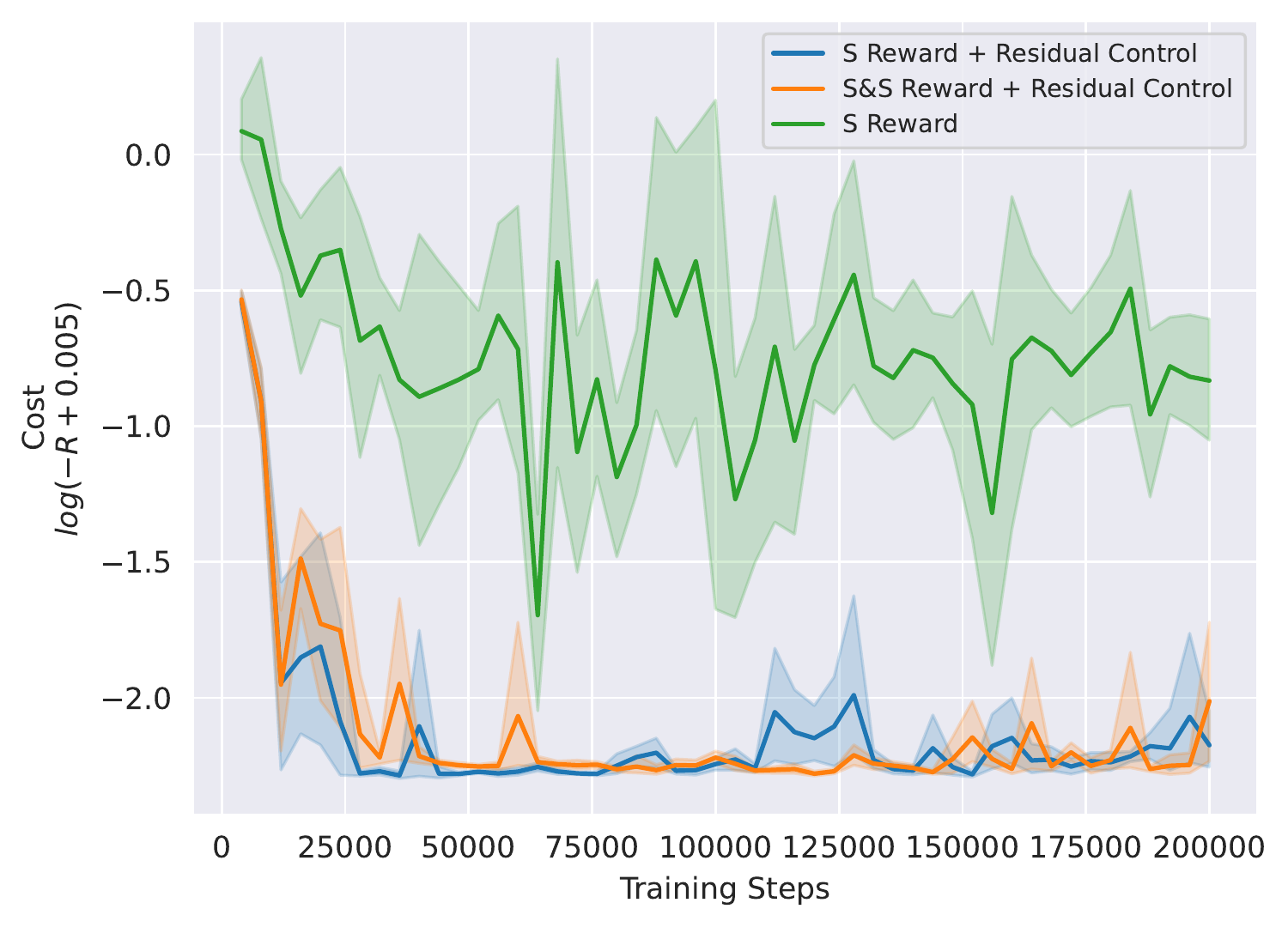}}
\vskip -3pt
\centering  
\subfigure[Critic loss during training]{\label{fig:b}\includegraphics[width=75mm]{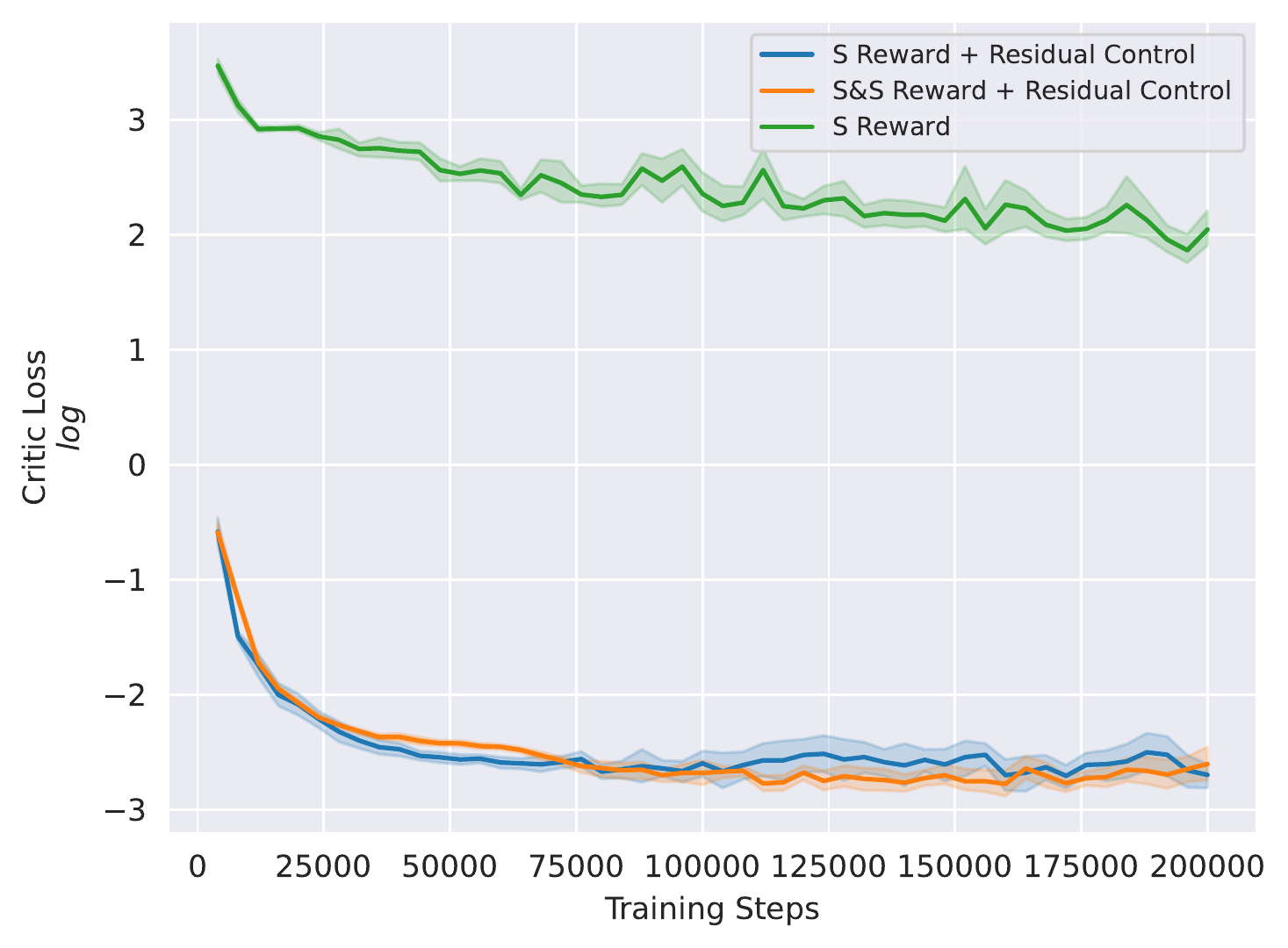}}
\caption{The plot illustrate the training progress with and w.o residual mechanism.}
\label{training}
\end{figure}

The DRL-controller is constructed using \textit{Multi-layer-perception} (MLP) that maps states to continuous actions. As shown in \ref{df}, the DRL-controller works together with the model-based controller 
to form the terminal control command $\mathbf{a}(k)$ as in~\eqref{residual}. For training, we take the cart-pole simulation provided in Open-AI gym~\cite{OpenAI-gym} and adapt it to a more realistic system with frictions and continuous action space. We leverage an off-policy \textit{actor-critic} algorithm DDPG~\cite{lillicrap2015continuous} to train the DRL-controller with the reward proposed in \eqref{reward}.

In the first experiment, we compare the stability performance of the system controlled by the model-based controller and Phy-DRL controller. We initialize the inverted pendulum in the neighbourhood of the equilibrium and let these two controllers to control the system respectively. As shown in Figure~\ref{trajectory}, the model based-controller fails in stablizing the inverted pendulum and eventually goes out of the safety bound. The reason is that the model-based controller is derived from the linearized dynamic model without friction force, which contains large model mismatch compared to the dynamics during test. In contrast, the Phy-DRL controller can stabilize the pendulum robustly around the equilibrium, as the DRL agent learned to deal with the modeling uncertainties and compensate the weakness of the model-based controller.

In the second experiment, we showcase the influence of the model-based controller during training. We implement the stability (S) encouraging reward function derived in~\cite{westenbroek2022lyapunov} without residual mechanism as a baseline. As shown in Figure~\ref{training}, the training with the proposed reward~\eqref{reward}, stability and safety (S$\&$S) encouraging reward, using residual control converges significantly faster than the baseline. The similar effect can also observed in the training with S reward using residual control. Since the S$\&$S reward has the similar scale as S reward, they eventually converge at similar level after approximately forty thousand training steps.

\section{Appendix}
\subsection{Proof of Lemma \ref{preenvelope}} \label{apreenvelope}
The condition of set \eqref{aset2} is equivalent to 
\begin{align}
\left[\underline{\mathbf{v}} + \mathbf{v}\right]_{i} \le \left[{\mathbf{D}}\right]_{i,:}\mathbf{s} \le \left[\overline{\mathbf{v}} + \mathbf{v}\right]_{i}, ~~i \in \{1,2,\ldots, h\}. \label{aset2p1}
\end{align}

{\underline{Case One: If $\left[\underline{\mathbf{v}} + \mathbf{v}\right]_{i} > 0$}}, we obtain from \eqref{aset2p1} that $\left[\overline{\mathbf{v}} + \mathbf{v}\right]_{i} > 0$, such that the \eqref{aset2p1} can be rewritten  as 
\begin{align}
&\frac{\left[{\mathbf{D}}\right]_{i,:}\mathbf{s}}{\left[\overline{\mathbf{v}} + \mathbf{v}\right]_{i}} = \frac{\left[{\mathbf{D}}\right]_{i,:}\mathbf{s}}{[\overline{\Lambda}]_{i,i}} \le 1, \nonumber\\
&\text{and}~\frac{\left[{\mathbf{D}}\right]_{i,:}\mathbf{s}}{\left[\underline{\mathbf{v}} + \mathbf{v}\right]_{i}} = \frac{\left[{\mathbf{D}}\right]_{i,:}\mathbf{s}}{[\underline{\Lambda}]_{i,i}} \ge 1 = [{\mathbf{d}}]_{i},i \in \{1,\ldots, h\}, \label{aset2p2aa}
\end{align}
which is obtained via considering the second items of \eqref{compbadd1} and \eqref{compbadd2} and the first item of \eqref{compbadd3}.

{\underline{Case Two: If $\left[\overline{\mathbf{v}} + \mathbf{v}\right]_{i} < 0$}}, we obtain from \eqref{aset2p1} that $\left[\underline{\mathbf{v}} + \mathbf{v}\right]_{i} < 0$, such that the \eqref{aset2p1} is rewritten equivalently as 
\begin{align}
&\frac{\left[{\mathbf{D}}\right]_{i,:}\mathbf{s}}{\left[\underline{\mathbf{v}} + \mathbf{v}_{\sigma}\right]_{i}} = \frac{\left[{\mathbf{D}}\right]_{i,:}\mathbf{s}}{[\underline{\Lambda}]_{i,i}} \le 1, \nonumber\\
&\text{and}~~\frac{\left[{\mathbf{D}}\right]_{i,:}\mathbf{s}}{\left[\overline{\mathbf{v}} + \mathbf{v}\right]_{i}} = \frac{\left[{\mathbf{D}}\right]_{i,:}\mathbf{x}}{[\overline{\Lambda}]_{i,i}} \ge 1  = [{\mathbf{d}}]_{i},i \in \{1,\ldots, h\}, \label{aset2p2}
\end{align}
which is obtained via considering the third items of \eqref{compbadd1} and \eqref{compbadd2} and the second item of \eqref{compbadd3}. 

{\underline{Case Three: If $\left[\overline{\mathbf{v}} + \mathbf{v}\right]_{i} > 0$ and $\left[\underline{\mathbf{v}} + \mathbf{v}\right]_{i} < 0$}}, the \eqref{aset2p1} can be rewritten equivalently as 
\begin{align}
&\frac{\left[{\mathbf{D}}\right]_{i,:}\mathbf{s}}{\left[\overline{\mathbf{v}} + \mathbf{v}\right]_{i}} = \frac{\left[{\mathbf{D}}\right]_{i,:}\mathbf{s}}{[\overline{\Lambda}]_{i,i}} \le 1, ~\text{and}\nonumber\\
&\frac{\left[{\mathbf{D}}\right]_{i,:}\mathbf{s}}{\left[-\underline{\mathbf{v}} - \mathbf{v}\right]_{i}} = \frac{\left[{\mathbf{D}}\right]_{i,:}\mathbf{s}}{[\underline{\Lambda}]_{i,i}} \ge -1 
 = [{\mathbf{d}}]_{i},i \in \{1,\ldots, h\}, \label{aset2p3}
\end{align}
which is obtained via considering the fourth items of \eqref{compbadd1} and \eqref{compbadd2} and the third item of \eqref{compbadd3}.

Finally, we note from the first items of \eqref{compbadd1} and \eqref{compbadd2} that the defined $\overline{\Lambda}$ and $\underline{\Lambda}$ are diagonal matrices. The conjunctive results \eqref{aset2p2aa}--\eqref{aset2p3} can thus be equivalent described by  
\begin{align}
\frac{{\mathbf{D}}\mathbf{x}}{\overline{\Lambda}} \le \mathbf{1}_{h}~~\text{and}~~\frac{{\mathbf{D}}\mathbf{x}}{\underline{\Lambda}} \ge {\mathbf{d}} \ge -\mathbf{1}_{h}, \nonumber
\end{align}
which is also equivalent to \eqref{aset2p1}. With the consideration of $\overline{\mathbf{D}} = \frac{{\mathbf{D}}}{\overline{\Lambda}}$ and $\underline{\mathbf{D}} = \frac{{\mathbf{D}}}{\underline{\Lambda}}$, we therefore conclude the statement, which completes the proof.

\bibliographystyle{ieeetr}
\bibliography{ref}

\end{document}